\documentclass[12pt]{article}

\usepackage{tikz}
\usetikzlibrary{shapes}
\usetikzlibrary{decorations.pathmorphing,decorations.pathreplacing,decorations.shapes}
\tikzset{every picture/.style={line width=0.75pt}} 
\usepackage{graphicx}

\usepackage{amssymb,amsmath,amsfonts}
\usepackage{hyperref}

\usepackage[margin=2cm]{geometry}
\usepackage{amsfonts,amsmath,amssymb}
\usepackage{fancyhdr}
\usepackage{graphicx, cleveref}

\usepackage{float}
\usepackage{url}
\usepackage[nottoc,notlot,notlof]{tocbibind}
\usepackage{color}
\usepackage{enumerate}
\usepackage{algorithm,algpseudocode}
\usepackage{multirow}
\usepackage{lscape}
\usepackage{multirow}
\usepackage{caption}
\usepackage{booktabs}
\usepackage{threeparttable}
\usepackage[rightcaption]{sidecap}
\usepackage{graphicx,wrapfig,lipsum}
\usepackage{array}%

\usepackage{authblk}
\usepackage{pict2e}

\usepackage{pgfkeys}
\usepackage{xcolor}

\usepackage{subcaption}
\usepackage{pgfgantt}
\usetikzlibrary{shadows}
\usetikzlibrary{shadings}

\usepackage{tikz}
\usetikzlibrary{mindmap}
\usetikzlibrary{shapes.geometric, arrows}

\usepackage{verbatim}
\usepackage{pgfplots}
\pgfplotsset{compat=1.18}

\usepackage{etoolbox}

\usetikzlibrary{shapes.geometric}

\usepackage{pstricks}
\usepackage{longtable}

\date{\normalsize \today}

\pgfplotsset{compat=1.18}
\begin{document}

\title{Early predicting of hospital admission using machine learning algorithms: Priority queues approach}


\author{
Jakub Antczak\thanks{
Wroc{\l}aw University of Science and Technology,
Wroc{\l}aw, Poland, email: 268745@student.pwr.edu.pl}
\ \
James Montgomery\thanks{University of Tasmania, Hobart, TAS 7001, Australia, email: james.montgomery@utas.edu.au}
\ \
Ma{\l}gorzata M. O'Reilly\thanks{University of Tasmania, Hobart, TAS 7001, Australia, email: malgorzata.oreilly@utas.edu.au}
\\
Zbigniew Palmowski\thanks{Wroc{\l}aw University of Science and Technology,
Wroc{\l}aw, Poland, email:
zbigniew.palmowski@pwr.edu.pl}
\ \
Richard Turner\thanks{School of Medicine, University of Tasmania, Australia, email: Richard.Turner@utas.edu.au}
}

\maketitle

\begin{abstract}
Emergency Department overcrowding is a critical issue that compromises patient safety and operational efficiency, necessitating accurate demand forecasting for effective resource allocation. This study evaluates and compares three distinct predictive models—Seasonal AutoRegressive Integrated Moving Average with eXogenous regressors (SARIMAX), EXtreme Gradient Boosting (XGBoost) and Long Short-Term Memory (LSTM) networks for forecasting daily ED arrivals over a seven-day horizon. Utilizing data from an Australian tertiary referral hospital spanning January 2017 to December 2021, this research distinguishes itself by decomposing demand into eight specific ward categories and stratifying patients by clinical complexity. To address data distortions caused by the COVID-19 pandemic, the study employs the Prophet model to generate synthetic counterfactual values for the anomalous period. Experimental results demonstrate that all three proposed models consistently outperform a seasonal naive baseline. XGBoost demonstrated the highest accuracy for predicting total daily admissions with a Mean Absolute Error of 6.63, while the statistical SARIMAX model proved marginally superior for forecasting major complexity cases with an MAE of 3.77. The study concludes that while these techniques successfully reproduce regular day-to-day patterns, they share a common limitation in underestimating sudden, infrequent surges in patient volume.
\end{abstract}



\newpage
\section{Highlights}
\begin{itemize}
     \item \textbf{Forecasting of ward-specific demand} --- we decompose demand into eight distinct hospital departments: Emergency Medicine, General Medicine, Surgery, Paediatric, Psychiatry, Cardiology, Neurology, and Other. This allows for targeted resource allocation that aligns with the siloed nature of hospital wards,
    \item \textbf{Stratification by clinical complexity} --- we stratify demand according to DRG complexity levels. Distinguishing high-complexity cases provides bed managers with a more robust metric for workload than raw patient counts, enhancing the efficiency of capacity planning and allocation \cite{OReilly20251493},
    \item \textbf{Identification of forecast difficulty} --- by splitting the data into wards, we can distinguish between patient groups that follow predictable trends and those that are difficult to predict.
 \end{itemize}

\newpage

\section{Introduction}
The modern healthcare landscape is defined by an enduring tension between the very high demand for acute medical services and the finite capacity of hospital systems to provide them. At the epicenter of this tension stands the Emergency Department (ED), which almost always suffers from overcrowding \cite{babatabar2020overcrowding}. Emergency department overcrowding occurs when patient volume exceeds the facility's capacity, leading to a disparity between supply and demand. This saturation negatively impacts operational performance as an excessive number of individuals wait for essential services\cite{lindner2021emergency}.

On a daily basis, tertiary referral hospitals are obliged to accept highly variable numbers of patients requiring acute care with varying degrees of complexity. Inability to accommodate and treat these arrivals in a timely fashion, a situation known as access block \cite{forero2011access}, can lead to adverse consequences for both patients \cite{bernstein2009effect} and the health workers required to make high-stakes decisions to priorities care in capacity-limited settings. Accurate forecasting of acute hospital arrivals over a defined time horizon is therefore crucial to enabling safe and efficient resource allocation.

However, existing literature predominantly focuses on forecasting total ED volume, which often lacks the granularity required for operational decision-making. Hospitals operate as a collection of distinct departments, and aggregate forecasts can mask significant variations in specific service lines.
This limitation is crucial, because proactive resource allocation requires more than raw admission predictions. Research by Lee at al., demonstrates that distinguishing between specific disposition destinations, such as intensive care or general practice units, yields significantly more actionable information for reducing boarding delays than aggregate models \cite{lee2020prediction}.

To address this, our research distinguishes itself by decomposing demand into eight specific ward categories: Emergency Medicine, General Medicine, Surgery, Paediatric, Psychiatry, Cardiology, Neurology, and Other. This approach facilitates targeted resource allocation that aligns with the siloed nature of hospital wards.

Furthermore, raw patient counts do not fully reflect the operational needs of the hospital. A patient with major clinical complexity requires significantly different resources than minor cases. Consequently, this study stratifies demand according to Diagnosis-Related Group (DRG) complexity levels. Distinguishing high-complexity cases provides bed managers with a more robust metric for workload than raw patient counts, enhancing the efficiency of capacity planning.

In this paper, we evaluate and compare three distinct predictive models: Seasonal AutoRegressive Integrated Moving Average with exogenous regressors (SARIMAX); EXtreme Gradient Boosting (XGBoost); and Long Short-Term Memory (LSTM) networks for forecasting daily ED arrivals over a seven-day horizon. The study utilizes data from an Australian tertiary referral hospital spanning January 2017 to December 2021. To address the significant data distortions caused by the COVID-19 pandemic during this timeframe, we employ the Prophet model to generate synthetic counterfactual values, ensuring our models learn from consistent demand patterns rather than anomalous pandemic fluctuations.

\section{Literature review}
The drive for accurate forecasting comes from from the need to mitigate ED overcrowding. Overcrowding is not merely a logistical inconvenience -- it fundamentally compromises patient safety and hospital efficiency. Research indicates that saturation leads to increased waiting times, causing patients to leave without being seen \cite{marino2024investigation}. This inability to access timely care can result in adverse outcomes, including higher mortality rates \cite{richardson2006increase} and increased readmissions due to premature discharge or inadequate treatment planning \cite{sartini2022overcrowding}. Consequently, researchers have employed numerous machine learning and statistical techniques to predict arrivals over short to medium term horizons. These predictive capabilities support data driven decision making regarding resource allocation and capacity planning. Specifically, the utility of time series models in forecasting ED arrivals has been widely investigated, yielding positive results across various and forecasting windows \cite{erkamp2021predicting, sudarshan2021performance, vieira2023forecasting, whitt2019forecasting}. Other studies have concentrated on shorter prediction horizons, specifically forecasting hourly patient volumes in the emergency department \cite{cheng2021forecasting, zhang2022forecasting}.

To improve forecast accuracy, recent literature incorporate exogenous variables. Calendar variables have emerged as some of the most significant predictors. Studies analyzing clinical data in Dutch and Danish hospitals found that factors such as the day of the week, month of the year, and public holidays significantly influence arrival frequencies \cite{erkamp2021predicting, sudarshan2021performance}. In Erkamp, Van Dalen and De Vries~\cite{erkamp2021predicting}, it was demonstrated that models relying solely on calendar variables could achieve high accuracy, noting distinct variations during holidays and weekends. Meteorological data is another investigated external driver. Some researchers have found strong correlations between specific weather conditions and ED presentations \cite{alvarez2024evaluating, erkamp2021predicting, pelaez2024explainable, sudarshan2021performance, zhao2022deep}. In Sch\"{a}fer, Walther, Grimm and H\"{u}bner~\cite{schafer2023combining}, they validated the necessity of department-specific feature analysis when integrating exogenous variables. In their study on optimizing patient-bed assignment, they observed that while trauma surgery admissions were strongly correlated with low temperatures and snow, other specialties like gastroenterology showed minimal weather sensitivity.

The methodology for predicting ED demand has evolved from classical statistical methods to advanced machine learning (ML) and deep learning (DL) techniques.

\subsection*{Statistical models}

In Zhang~et~al.~\cite{zhang2022forecasting} the ARIMA model was compared against several ensemble learning algorithms, including XGBoost, Random Forest, Gradient Boost, and AdaBoost. The analysis revealed that ARIMA yielded significantly higher RMSE and MAE values. Consequently, the study concluded that the ensemble methods outperformed ARIMA in forecasting hourly patient arrivals. Limitations of the standard ARIMA model were further evidenced by Gafni-Pappas and Khan~\cite{gafni2023predicting}, who reported that it produced the highest RMSE among the models tested on daily visit data. To address the constraints of univariate forecasting, Tuominen~et~al.~\cite{tuominen2022forecasting} employed ARIMAX to integrate exogenous variables into the dataset. Specifically, they developed an ARIMAX-W model trained on weather and calendar data, which facilitated multistep forecasting for specific horizons. This approach allows the model to predict next day arrivals by synthesizing prior forecasts with input data, while assuming zero unobserved residuals. To address the non-seasonal limitations of standard ARIMA, several studies have adopted the Seasonal Autoregressive Integrated Moving Average (SARIMA) model to capture periodic patterns in ED arrivals \cite{vieira2023forecasting, tuominen2022forecasting}. However, computational constraints have been noted. Both Rocha and Rodriguez~\cite{rocha2021forecasting} and Vieira, Sousa and D\'{o}ria-N\'{o}bregaof~\cite{vieira2023forecasting} mentioned a software limit of 350 seasonal periods. To mitigate this restriction, Rocha and Rodriguez~\cite{rocha2021forecasting} implemented a weekly seasonality structure. Building on the capabilities of SARIMA, recent research has focused on the SARIMAX model, which integrates external variables to enhance predictive accuracy. Whitt and Zhang~\cite{whitt2019forecasting} demonstrated that including external regressors significantly improved performance for daily arrival forecasts, achieving a lower MSE compared to standard SARIMA. This advantage extends to granular, short-term predictions. Cheng~et~al.~\cite{cheng2021forecasting} utilized SARIMAX with operational variables to forecast hourly arrivals up to four hours ahead, yielding the best performance metrics among tested models.

\subsection*{Machine learning models}

In the context of ED forecasting, XGBoost has demonstrated superior performance for short-term horizons. Rocha and Rodriguez~\cite{rocha2021forecasting} utilized Bayesian optimization to minimize mean absolute error, finding that while XGBoost required over 3000 iterations, it emerged as the most efficient model for hourly presentations, ranking second only to SARIMA in accuracy. Similarly, in both King~et~al.~\cite{king2022machine} and Zhang~et~al.~\cite{zhang2022forecasting}, XGBoost was identified as the best approach for predicting hourly arrivals. Also, King~et~al.~\cite{king2022machine} highlighted the model's utility in feature selection, noting its ability to identify critical variables within complex datasets. Furthermore, the predictive capabilities of XGBoost can be significantly augmented by integrating non-traditional data sources. Seo~et~al.~\cite{seo2024prediction} demonstrated that incorporating unstructured text data from clinical notes improved the model's AUROC by approximately 6\% for predicting hospitalization within 24 hours. Beyond gradient boosting methods, Random Forest (RF) has been widely implemented in ED forecasting due to its ability to prevent overfitting through the creation of multiple decision trees. As an ensemble learning method, it effectively handles both classification and regression problems by aggregating outputs from random observation samples. In practical applications for predicting ED arrivals up to seven days in advance, Random Forest models have shown that performance often peaks at around 100 trees, although increasing the number of trees comes with higher computational costs \cite{sudarshan2021performance}. Another classic method applied in this domain is Na\"ive Bayes \cite{ali2023extreme, hong2018predicting}, which utilizes conditional probability calculations suitable for time forecasting. In Kishore~et~al.~\cite{kishore2023early}, Na\"ive Bayes classifiers have achieved accuracy rates between 84\%--88\% in predicting patient arrivals. Harrou~et~al.~\cite{harrou2020forecasting} observed Na\"ive Bayes classifiers performed particularly well when predicting a patient's length of stay, sometimes outperforming other predictive models in that specific metric.

\subsection*{Deep learning models}

Deep learning techniques have increasingly replaced traditional statistical methods for Emergency Department (ED) forecasting due to their ability to model complex non-linear relationships and high-dimensional data. Among these, Long Short-Term Memory (LSTM) networks are the most widely applied deep learning architecture for time-series problems. Zhao~et~al.~\cite{zhao2022predicting} demonstrated that deep architectures, specifically those with 2 to 3 layers and 200--300 hidden units, achieved the best prediction performance for daily ED visits, outperforming shallower Bidirectional LSTM (BiLSTM) and Convolutional LSTM (ConvLSTM) models configured with only a single layer. Similarly, other studies have successfully employed multi-layered LSTM and BiLSTM models \cite{harrou2020forecasting, zhao2022deep}. While LSTMs excel at capturing temporal dependencies, Convolutional Neural Networks (CNNs) are often integrated to enhance feature extraction \cite{zhao2022predicting}. Research comparing LSTMs with CNNs has led to the development of hybrid models that leverage the strengths of both. For instance, Harrou~et~al.~\cite{harrou2022effective} proposed an LSTM-CNN hybrid to predict patient flow. In this architecture, 1D Convolutional layers (Conv1d) with ReLU activation and Max Pooling are used to extract local invariant features from the input data, which are subsequently processed by LSTM layers to model the temporal evolution. The efficacy of these deep learning models is frequently enhanced through multivariate analysis and rigorous hyperparameter optimization. Etu~et~al.~\cite{etu2022comparison} found that Multivariable LSTM models, which incorporate external meteorological variables (e.g., temperature, humidity, precipitation), significantly outperformed univariate counterparts and traditional models like SARIMAX. To maximize performance, researchers advocate for the use of Grid Search strategies to finetune weights and biases.

\subsection*{Simulation and queuing models}

While this study focuses on accurate demand forecasting, these predictions are ultimately intended to inform capacity management systems, often modeled as queues. Koizumi, Kuno and Smith~\cite{koizumi2005modeling} established the foundational importance of queuing networks with blocking for modeling congestion and access block in healthcare systems. Recently, the field has advanced toward generative simulation. Holt~et~al.~\cite{Holt202523513} introduced G-Sim, a framework that utilizes LLMs to automate the construction and calibration of these simulators. This progression highlights the growing intersection between predictive modeling and dynamic simulation environments for resource allocation.

Despite the breadth of methodologies applied to total patient volumes, there remains a shortage of research focused on decomposing demand by specific hospital wards or clinical complexity. Most existing studies forecast aggregate arrivals, yet resource allocation often requires granular data, such as predicting specific demands for surgery, paediatrics, or psychiatry. This study aims to bridge this gap by comparing statistical, ML, and DL approaches to forecast ward-specific demand stratified by clinical complexity, thereby offering more actionable insights for hospital management.

\section{Dataset}
Our dataset consists of Emergency Department (ED) patient arrival records from January 1, 2017 through December 31, 2021 at an Australian tertiary referral hospital. Each patient entry includes the date of ED arrival, the ward the patient was admitted, and their diagnosis-related group (DRG), which is a code indicating the case’s complexity and category. Although DRG codes are administrative labels finalized at discharge, they are strongly correlated with the preliminary diagnoses and acuity assessments available shortly after arrival. Therefore, we treat the DRG as a representative label for the clinical information available early in the patient stay \cite{liu2021early}. We aggregated the data into daily counts of arrivals and classified them into eight categories according to the ward of admission: Emergency Medicine, General Medicine, Surgery, Paediatric, Psychiatry, Cardiology, Neurology, and Other. This resulted in 1826 daily observations for each of the main categories. For each ward category we created a subset based on complexity \cite{IHACPA2022}. We have three types of complexity in our dataset: major, intermediate and minor. We focused on major complexity cases and  complexity intermediate and minor cases were aggregated into other. This process led us to constructing 16 daily time series (Emergency Medicine Major, Emergency Medicine Other, etc.) each representing the number of arrivals of that type per day.

\begin{figure}[h]
    \centering
    \includegraphics[width=\linewidth]{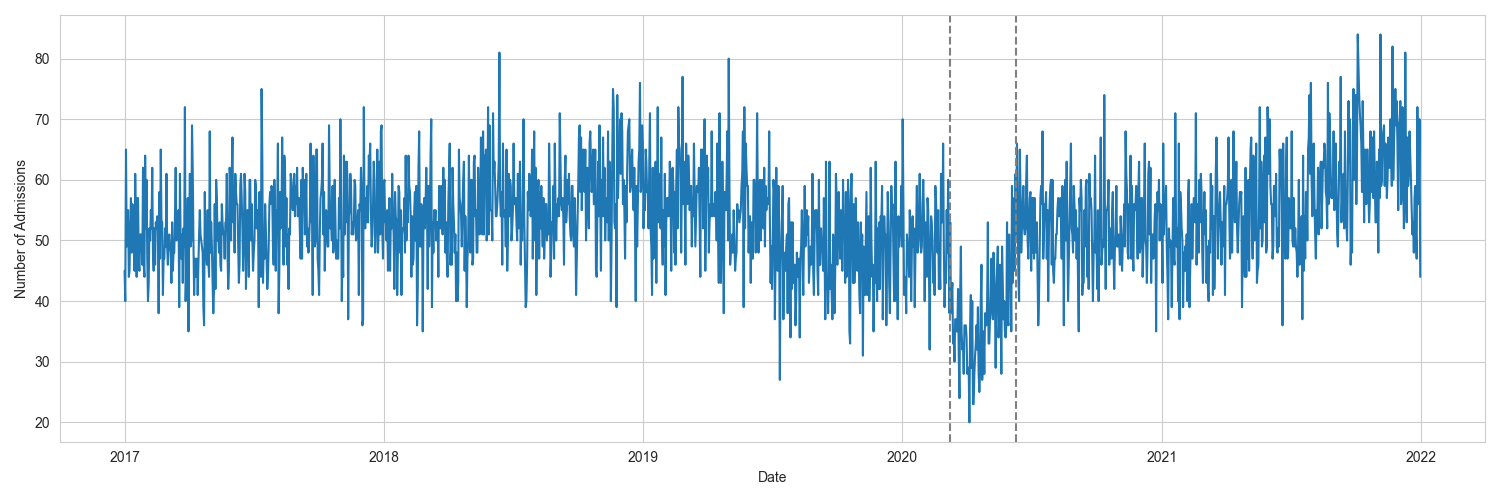}
    \caption{Daily number of admissions from 2017-01-01 to 2021-12-31, with the COVID-19 period indicated by vertical lines. }
    \label{fig1:daily_admissions_all}
\end{figure}

During the early stages of the COVID-19 pandemic, rapidly changing public health restrictions significantly affected ED patient admissions. As shown in Figure~\ref{fig1:daily_admissions_all}, the period corresponding to these restrictions (between the two vertical dashed lines) is characterized by an enormous decrease in the total number of arrivals. These extreme fluctuations tend to distort parameter estimation and degrade the predictive performance of standard time–series models. To mitigate this issue, we treated this interval as an anomalous period and replaced the observed counts with counterfactual values obtained from the Prophet model, a decomposable time-series forecasting method developed by Meta. Prophet is a decomposable time-series forecasting model that formulates the prediction as an additive regression of three main components - a trend, seasonality and holiday effects. We selected Prophet for this imputation task because its additive structure is particularly effective at isolating and reproducing the strong seasonal patterns and calendar effects inherent in hospital demand, allowing for a robust estimation of  trajectories without COVID.

\begin{figure}[h]
    \centering
    \includegraphics[width=\linewidth]{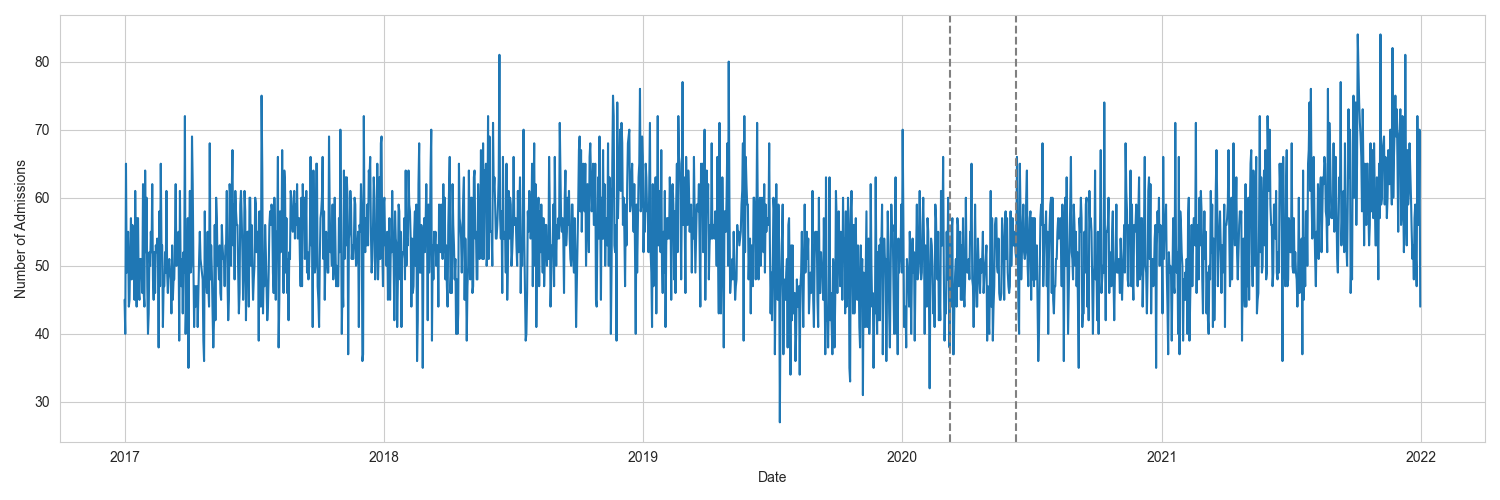}
    \caption{Daily number of admissions from 2017-01-01 to 2021-12-31, with the COVID-19 period replaced with synthetic data.}
    \label{fig2:fig_admissions_all_synth}
\end{figure}

For each of the 16 daily arrival series we fitted a Prophet model using only the non-COVID data. The fitted model was then used to produce forecasts $\hat{y}_t$ for all days in the study period. We estimated the standard deviation $\sigma$ of the in-samples residuals $y_t - \hat{y}_t$ on the training set and, for dates within the COVID window, generated stochastic synthetic counts
$$y_t^* =\max{\{0, \texttt{round}(\hat{y}_t + \epsilon_t)\}},$$
where $\epsilon_t\sim \mathcal{N}(0,\sigma^2)$. This replaced the observed values for the COVID period, which can be observed on the Figure \ref{fig2:fig_admissions_all_synth}.

In addition to the admission records, we linked our data with daily meteorological observations for the region where the hospital is located. The weather dataset contains calendar date, air temperature, wind speed and precipitation. For each date we calculated the daily maximum and minimum temperature, as well as the mean wind speed. These summaries were then merged with the daily admissions table. We also added a column indicating if the date was a local public holiday.

\begin{table}[h]
\centering
\resizebox{\linewidth}{!}{%
\begin{tabular}{lccc}
\hline
\multicolumn{1}{c}{\textbf{Ward}} & \textbf{Number of Admissions} & \textbf{Major Complexity} & \textbf{Other Complexity} \\ \hline
\textbf{Emergency Medicine}       & 15.18                         & 2.35                      & 12.82                     \\
\textbf{General Medicine}         & 11.28                         & 5.92                      & 5.38                      \\
\textbf{Surgery}                  & 10.11                         & 2.58                      & 7.52                      \\
\textbf{Paediatric}               & 2.82                          & 0.85                      & 2.46                      \\
\textbf{Psychiatry}               & 3.20                          & 0.66                      & 2.17                      \\
\textbf{Cardiology}               & 2.52                          & 0.78                      & 1.75                      \\
\textbf{Neurology}                & 0.68                          & 0.24                      & 0.45                      \\
\textbf{Other}                    & 7.83                          & 3.00                      & 4.83                      \\
\textbf{Total Arrivals}           & 53.70                         & 16.32                     & 37.38
\end{tabular}%
}
\caption{Average patient arrivals in each ward by major \& other complexity.}
\label{tab1:average_patients_arrivals}
\end{table}

From Table \ref{tab1:average_patients_arrivals}, we observe that emergency medicine typically has the highest volume of daily admissions, whereas specialized categories like neurology or cardiology have fewer daily cases on average. In terms of complexity, major complexity cases form a smaller fraction of arrivals in each category, which is expected since high-complexity cases are less common than routine.

\section{Methods}
\label{sec:methods}

Our forecasting task is defined as follows: \textbf{given the past 14 days of data, predict the next 7 days of daily arrivals}. This 14-day input window provides recent trend and weekly pattern information, and the 7-day forecast horizon produces a full week of predictions to aid short-term planning. We consider three modelling approaches: a statistical time-series model, SARIMAX, and two machine-learning models, XGBoost and LSTM. For XGBoost and LSTM, the problem is formulated as supervised learning on sliding 14-day windows of the historical time series, directly producing 7-day multistep forecasts. In contrast, SARIMAX models the temporal dependence structure of the series together with exogenous variables, and is used to generate the same 7-day-ahead forecasts.

\subsection{SARIMAX}
\label{sec:sarimax}

We employ a Seasonal AutoRegressive Integrated Moving Average with eXogenous regressors (SARIMAX) model to capture temporal dependence together with the effect of external covariates. For each series, let $y_t$ denote the daily arrivals. The model is specified as
\begin{equation*}
    \varphi_p(B)\Phi_P(B^s)\nabla^d \nabla^D_s y_t = \beta_k \mathbf{X}_{k,t}^\text{T} + \theta_q(B)\Theta_Q(B^s)\epsilon_t,
\end{equation*}
where $\varphi_p(B)$ is the regular autoregressive (AR) polynomial of order $p$, $\Phi_P(B^s)$ is the seasonal AR polynomial of order $P$, $\nabla^d$ is differencing operator, $\nabla^D_s$ is seasonal differencing operator, $\theta_q(B)$ is the regular moving average (MA) polynomial of order $q$, $\Theta_Q(B^s)$ is the seasonal MA polynomial of order $Q$, $B$ is the back-shift operator, which operates on the observation $y_t$ shifting it one point in time (i.e., $B^k(y_t)=y_{t-k}$), $\epsilon_t$ is white noise process, $s$ is the seasonal period, $\mathbf{X}_{k,t}$ is the vector including the $k^{th}$ explanatory input variables at time $t$ and $\beta_k$ is the coefficient value of the $k^{th}$ exogenous input variable. All operators and polynomial are defined as
\begin{align*}
\varphi_p(B) &= 1 - \sum_{i=1}^{p} \varphi_i B^{i},
& \Phi_P(B^{s}) &= 1 - \sum_{i=1}^{P} \Phi_i B^{s,i}, \\
\theta_q(B) &= 1 - \sum_{i=1}^{q} \theta_i B^{i},
& \Theta_Q(B^{s}) &= 1 - \sum_{i=1}^{Q} \Theta_i B^{s,i}, \\
\nabla^{d} &= (1-B)^{d},
& \nabla^{D}_{s} &= (1-B^{s})^{D}.
\end{align*}
More details can be found in Box~et~al.~\cite{box2015time}.

\subsection{XGBoost}
\label{sec:xgboost}

EXtreme Gradient Boosting (XGBoost) is a highly efficient and scalable implementation of the gradient boosting framework. It functions as an ensemble learning method that constructs a predictive model by combining the outputs of multiple weak learners, specifically regression trees, in a sequential manner to minimize a regularized objective function \cite{DBLP:journals/corr/ChenG16}. For each target series we cast multistep forecasting as a supervised learning problem on overlapping input–output pairs. For a given day $t$, let $y_t$ denote the daily number of admissions for this series and let $\mathbf{Z}_t \in \mathbb{R}^{L \times F}$ denote the feature matrix containing $F$ daily features over the last $L=14$ days. For XGBoost, $\mathbf{Z}_t$ is flattened into a single feature vector $\mathbf{X}_t\in \mathbb{R}^{L \cdot F}$. The associated target is the vector of admissions over the next $H=7$ days. We adopt a direct multistep strategy and learn a separate gradient-boosted tree ensemble for each forecast horizon $h=1,\dots,H$. For horizon $h$, the prediction is
$$\hat{y}_{t+h}=f_h(\mathbf{X}_t)=\sum_{m=1}^Mf_{m,h}(\mathbf{X}_t),$$
where each $f_{m,h}$ is a regression tree and $M$ is the number of boosting iterations. The ensembles are trained independently to minimize the squared-error loss with $\ell_2$ regularization.

\subsection{LSTM}
\label{sec:lstm}

Long Short-Term Memory networks (LSTMs) represent a subclass of Recurrent Neural Networks (RNNs). They were specifically developed to mitigate the vanishing gradient issue prevalent in standard RNNs, thereby allowing the model to effectively capture and preserve long-range temporal dependencies with sequential datasets \cite{hochreiter1997long}. Similar to the XGBoost approach, we formulate the problem using a sliding window of length $L=14$. For a given day $t$, let $\mathbf{X}_t \in \mathbb{R}^{L \times F}$ denote the sequence of feature vectors over the last $L$ days, where $F$ is the number of input features. Unlike the tree-based approach, we do not flatten the feature matrix, instead the standardized sequence is fed directly into the network. Our architecture consists of a stacked LSTM with $2$ recurrent layers and a hidden state size of $32$ units. The network processes the input sequence $\mathbf{X}_t$ and extracts the final hidden state $h_t$, which summarizes the temporal context up to time $t$. This representation is passed through a dropout layer with a probability of $0.2$ to prevent overfitting, followed by a fully connected block consisting of a linear layer with 128 units and ReLU activation function. Finally, a projection layer maps the output to a vector of size $H=7$. Consequently, the model performs direct multistep forecasting, producing the entire 7-day forecast vector $\hat{y}_{t+1:t+H}$ in a single forward pass
$$\hat{y}_{t+1:t+H}=\mathcal{\psi}(h_t;W),$$
where $\mathcal{\psi}$ represents the dense layers and $W$ denotes the learnable weights of the network. The model is trained to minimize the mean squeared error loss using the Adam optimizer.
\begin{figure}[h]
    \centering
    \includegraphics[width=\linewidth]{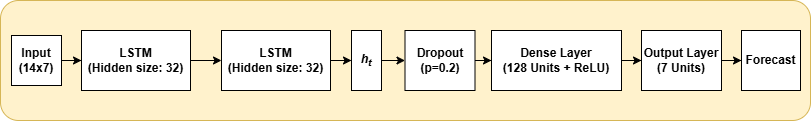}
    \caption{Architecture of the proposed LSTM model for multistep forecasting.}
    \label{fig:architecture}
\end{figure}

\section{Experiments and results}
\label{sec:experiments}

For each target series we first construct a feature set that combines meteorological, calendar, and autoregressive information. We include calendar effects (day of week, month, day of year, weekend flag) and a cyclic encoding of day of week. To capture temporal dependence in each target, we add lags of 1--7 days and rolling statistics: 7 day and 14 day moving means and standard deviations (except for SARIMAX, due to the model nature). We also compute days since last holiday. After removing initial rows with insufficient history for rolling windows, we sort the data chronologically and split it into a training/validation part and a hold-out test part, taking the last 180 days of the series as the test period (around 10\% of the total series). From these segments we then build supervised sequences. Each sample consists of 14 consecutive days of features as input and the corresponding 7 subsequent days of the target variable as output.

To select the hyperparameters for each model we performed a grid search. The search space for XGBoost included number of trees $\in \{ 100,300,500\}$, learning rate $\in \{0.01, 0.05, 0.1\}$ and maximum tree depth $\in \{3, 6, 9\}$. For LSTM it included hidden sizes $\in \{16,32,64,128\}$, number of recurrent layers $\in \{1,2,3\}$, dropout rate $\in \{0.1, 0.2, 0.3\}$, dense units $\in \{16,32,64,128\}$, learning rates $\in \{\in 0.0001, 0.001, 0.01\}$ and batch sizes $\in \{8,16,32,64,128\}$. For XGBoost the best parameter configuration was: number of trees $100$, learning rate $0.05$, maximum tree depth $3$, for the LSTM, the best configuration was: hidden size $32$, number of recurrent layers $2$, dropout rate $0.2$, fully connected layer $128$, learning rate $0.01$, batch size $64$. For the SARIMAX model, we selected the optimal configuration using the Akaike Information Criterion (AIC), which resulted in a non-seasonal order of $p=0$, $d=1$, $q=2$ and a seasonal order of $P=0$, $D=1$, $Q=1$, $s=7$. All searches were performed on the time series of total arrivals.

After fixing the hyperparameters, all three models were trained separately for each ward and target series. To evaluate the stability of the machine learning approaches and account for stochasticity in initialization and optimization, we performed 10 independent training runs for both XGBoost and LSTM, using a distinct random seed for each iteration. For these models, training examples were created from the training segment using the sliding-window scheme described above. XGBoost models were fitted on all training windows and produced 7-dimensional forecasts. For the LSTM network, input features and targets were standardized based on the training set. The model was trained with the Adam optimiser, using 20\% of the training windows as a validation set and early stopping with a patience of 10 epochs based on validation loss. In contrast, as SARIMAX is a deterministic statistical method, it was fitted once to the full training period of each series, using the selected orders and the exogenous covariates. We then generated one-step-ahead forecasts recursively over the 180-day test period and grouped them into 7-day horizons to ensure comparability with the XGBoost and LSTM forecasts. Consequently, the performance metrics for XGBoost and LSTM represent the aggregate performance (median and standard deviation) across the 10 runs,  SARIMAX results reflect its single  output.

We computed the Absolute Percentage Error (APE) for each forecast day as
\begin{equation*}
    \text{APE}_t = \begin{cases}
    \frac{|y_t - \hat{y}_t|}{|y_t|}, & y_t >0,\\
    \text{undefined}, & y_t=0,
    \end{cases}
\end{equation*}
and then averaged APE over all test days (and horizons) within a ward-model pair to obtain MAPE in percent. Thus, days with zero true arrivals were excluded from the percentage-error calculation to avoid division by zero. We also compared our results by Mean Absolute Error (MAE), which is $\text{MAE} = \frac{\sum_{i=1}^n|y_i - \hat{y}_i|}{n}$.

As a baseline, we evaluated a simple seasonal naive model. For each forecast horizon the baseline assumes that demand in a given day will be the same as on the corresponding day of the previous week, i.e.
$$\hat{y}_{t+h} = y_{t+h-7}, \quad h = 1,\dots,7.$$
This provides an intuitive comparison: any useful forecasting method should outperform a rule that simply repeats the value from one week ago.

\subsection{Admissions with major complexity}\label{admissions_major}
\begin{table}[h]
\centering
\resizebox{\textwidth}{!}{%
\begin{tabular}{lcccc|cccc}
\hline
                     & \multicolumn{4}{c|}{\textbf{MAE}}                                                       & \multicolumn{4}{c}{\textbf{MAPE {[}\%{]}}}                                                \\ \hline
\textbf{Wards/Model} & LSTM               & SARIMAX       & XGBoost            & Baseline & LSTM                & SARIMAX        & XGBoost             & Baseline \\ \hline
Emergency Medicine   & 1.84 $\pm$ 0.05          & 1.86          & \textbf{1.78 $\pm$ 0.02} & 1.93                         & \textbf{46.14 $\pm$ 0.57} & 46.45          & 46.38 $\pm$ 0.42          & 66.93                        \\
General Medicine     & 2.24 $\pm$ 0.08          & 2.24          & \textbf{2.20 $\pm$ 0.01} & 2.95                         & 38.96 $\pm$ 1.87          & 45.79          & \textbf{38.57 $\pm$ 0.25} & 54.40                        \\
Surgery              & 1.31 $\pm$ 0.03          & \textbf{1.28} & 1.30 $\pm$ 0.01          & 1.73                         & \textbf{47.86 $\pm$ 1.99} & 54.72          & 50.39 $\pm$ 0.52          & 69.75                        \\
Paediatric           & 0.83 $\pm$ 0.00          & 0.83          & 0.83 $\pm$ 0.00          & 1.21                         & \textbf{29.64 $\pm$ 0.00} & \textbf{29.64} & 30.27 $\pm$ 0.33          & 68.90                        \\
Psychiatry           & 0.77 $\pm$ 0.04          & \textbf{0.71} & 0.72 $\pm$ 0.01 & 1.04                         & 46.58 $\pm$ 14.68         & \textbf{23.93} & 30.18 $\pm$ 0.82          & 63.92                        \\
Cardiology           & 0.77 $\pm$ 0.01          & 0.77          & 0.77 0.00          & 0.88                         & 23.17 $\pm$ 5.81          & \textbf{20.65} & 21.75 $\pm$ 0.43          & 75.29                        \\
Neurology            & 0.30 $\pm$ 0.00          & \textbf{0.29} & 0.31 $\pm$ 0.00          & 0.44                         & 100.00 $\pm$ 0.00         & 100.00         & 99.52 $\pm$ 0.37          & \textbf{78.31}               \\
Other                & \textbf{1.36 $\pm$ 0.02} & \textbf{1.36} & 1.37 $\pm$ 0.01          & 1.90                         & 54.82 $\pm$ 1.75          & \textbf{51.91} & 55.25 $\pm$ 0.39          & 75.47                        \\
Total Arrivals       & 4.49 $\pm$ 0.26          & \textbf{3.77} & 4.14 $\pm$ 0.02          & 5.47                         & 21.87 $\pm$ 0.82          & \textbf{20.07} & 20.60 $\pm$ 0.1          & 29.53    \\
\hline
\end{tabular}
}%
\caption{Forecasting performance for major complexity admissions by ward. Values for XGBoost and LSTM represent the mean $\pm$ standard deviation across 10 independent experimental runs.}
\label{tab2:forecasting_performance_major}
\end{table}

Table~\ref{tab2:forecasting_performance_major} reports the forecasting performance for major-complexity admissions by ward. Overall, the LSTM, SARIMAX and XGBoost models achieve broadly similar results across wards and outperform our baseline. At the ward level, MAE values typically lie between about 0.3 and 2.25 patients per day, meaning that on average all three methods miss by fewer than two major admissions per day in each specialty. In terms of MAPE, the models exhibit moderate relative errors in the larger groups: for Emergency Medicine and Surgery the typical error is around 46--55\%, while Paediatric, Psychiatry and Cardiology show lower MAPEs, with SARIMAX in particular substantially reducing error for Psychiatry. The Neurology group has a MAPE of $100\%$ for two models. This reflects the very small daily counts in this category. When the true value is $1$ or $2$ patients, an absolute error of just one patient already corresponds to a $50$–$100\%$ percentage error, therefore these high MAPEs should be interpreted with caution for low-volume wards. We can also see that Neurology is the only ward where the proposed models did not outperform the baseline in terms of MAPE.

When aggregating across all wards, the models perform better in relative terms. The total major complexity arrivals are predicted with an MAE of $4.49 \pm 0.26$ (LSTM), $3.77$ (SARIMAX) and $4.14 \pm 0.02$ (XGBoost), corresponding to MAPE values of $21.87\% \pm 0.82 \%$, $20.07\%$ and $20.60\% \pm 0.1 \%$, respectively. This indicates that, at the level of overall ED major admissions, SARIMAX provides the most accurate forecast, although the differences between the three approaches are modest and all capture the main temporal patterns in the data and outperform the baseline.

\begin{figure}[h]
    \centering
    \includegraphics[width=\linewidth]{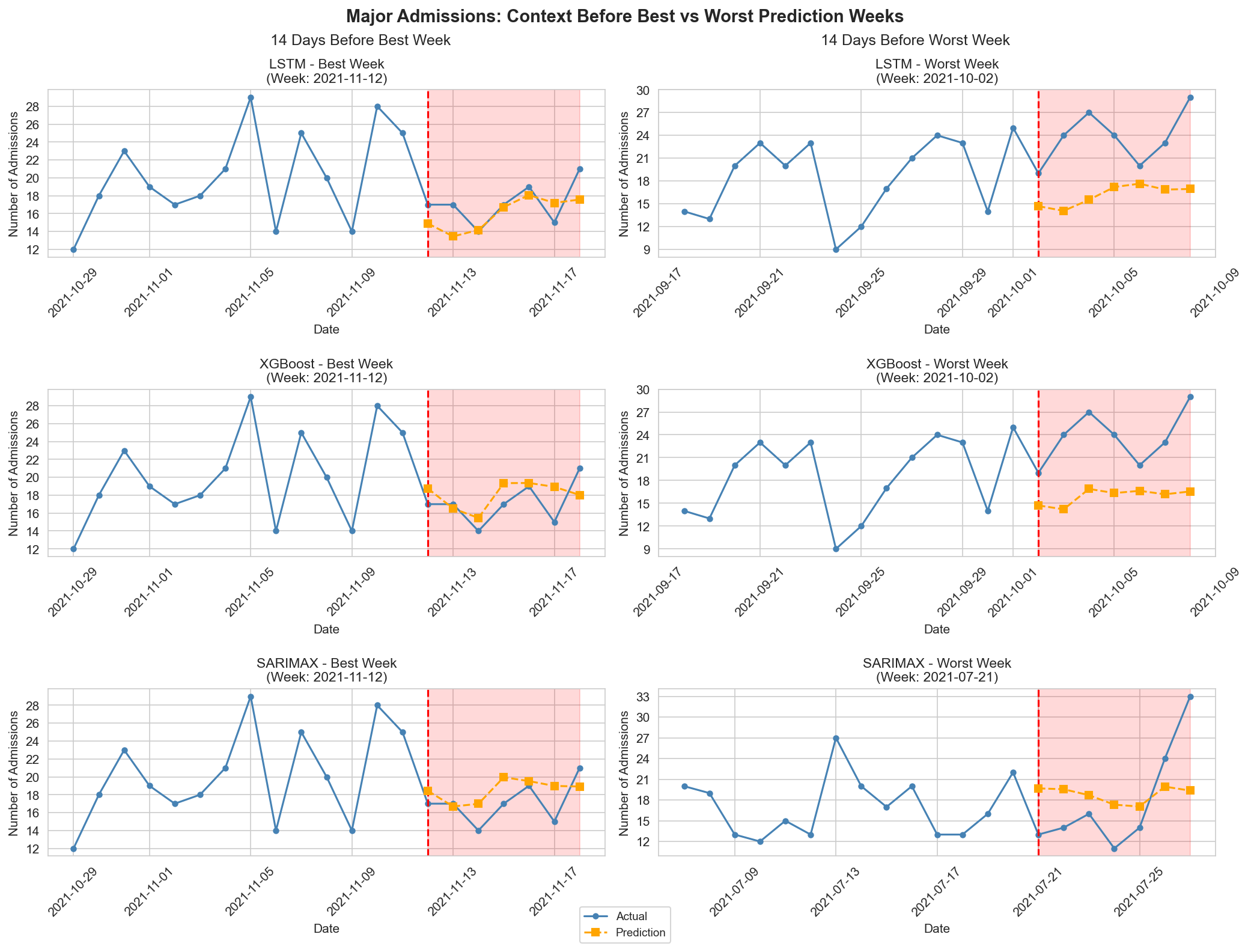}
    \caption{Actual and predicted total arrivals on best week and worst week for major complexity admissions -- comparison between XGBoost, LSTM and SARIMAX.}
    \label{fig:best_model_predictions_comparison_major}
\end{figure}

Figure \ref{fig:best_model_predictions_comparison_major} illustrates the model performance during the best and worst prediction weeks for major complexity admissions (for selected random seed). The left column displays the 7-day windows where each model achieved its lowest MAE. In each panel on the left, the predicted trajectories reproduce the observed day-to-day pattern. Each forecast tends to overestimate on the lowest-demand day. Overall, the models are able to closely reproduce the observed pattern of changes, with the LSTM and XGBoost achieving the lowest MAE of $1.14$. The right column of Figure \ref{fig:best_model_predictions_comparison_major} shows the worst forecast weeks for major complexity arrivals. The models substantially underestimate the number of arrivals. Among the models, SARIMAX performs best, with an MAE of $6.14$, whereas XGBoost performs worst, with an MAE of $7.71$. These examples highlight a common limitation of our approaches – the models handle regular weeks reasonably well, but struggle with rare surges in demand, tending to smooth them towards the recent average. Appendix~\ref{app:feature_importance} presents XGBoost's reported importance of input features, showing that the average number of recent arrivals is a strong contributor to its predictions.

\subsection{Overall admissions}
\label{sec:admissions}

\begin{table}
\centering
\resizebox{\textwidth}{!}{%
\begin{tabular}{lcccc|cccc}
\hline
                     & \multicolumn{4}{c|}{\textbf{MAE}}                                                    & \multicolumn{4}{|}{\textbf{MAPE {[}\%{]}}}                           \\ \hline
\textbf{Wards/Model} & LSTM               & SARIMAX       & XGBoost            & Baseline                  & LSTM                & SARIMAX        & XGBoost             & Baseline \\ \hline
Emergency Medicine   & 4.71 $\pm$ 0.3           & 7.14          & \textbf{4.38 $\pm$ 0.04} & 4.39 & 20.87 $\pm$ 1.34          & 32.38          & \textbf{19.51 $\pm$ 0.17} & 21.65    \\
General Medicine     & 2.93 $\pm$ 0.02          & \textbf{2.83} & 2.87 $\pm$ 0.01          & 3.60 & 29.02 $\pm$ 0.64          & 30.40          & \textbf{28.27 $\pm$ 0.09} & 34.81    \\
Surgery              & 2.60 $\pm$ 0.02          & \textbf{2.50} & 2.54 $\pm$ 0.01          & 3.73 & 29.65 $\pm$ 0.78          & 29.47          & \textbf{29.25 $\pm$ 0.12} & 41.57    \\
Paediatric           & 1.87 $\pm$ 0.04          & \textbf{1.80} & 1.90 $\pm$ 0.01          & 2.25 & 45.77 $\pm$ 1.0           & 52.11          & \textbf{45.41 $\pm$ 0.29} & 61.52    \\
Psychiatry           & \textbf{1.48 $\pm$ 0.01} & 1.52          & 1.49 $\pm$ 0.01          & 2.15 & \textbf{45.07 $\pm$ 0.96} & 57.41          & 48.12 $\pm$ 0.26          & 75.38    \\
Cardiology           & 1.36 $\pm$ 0.07          & 1.32          & \textbf{1.31 $\pm$ 0.01} & 1.56 & 63.91 $\pm$ 6.05          & 63.09          & \textbf{62.49 $\pm$ 0.49} & 70.49    \\
Neurology            & 0.65 $\pm$ 0.02          & \textbf{0.63} & 0.66 $\pm$ 0.01          & 0.75 & 27.93 $\pm$ 10.78         & \textbf{13.55} & 24.60 $\pm$ 1.19          & 63.16    \\
Other                & 2.17 $\pm$ 0.03          & 2.18          & \textbf{2.12 $\pm$ 0.01} & 2.97 & 36.18 $\pm$ 0.72          & 37.97          & \textbf{35.04 $\pm$ 0.14} & 46.04    \\
Total Arrivals       & 6.88 $\pm$ 0.21          & 7.28          & \textbf{6.63 $\pm$ 0.03} & 8.14 & 10.70 $\pm$ 0.27          & 11.18          & \textbf{10.32 $\pm$ 0.04} & 13.33    \\
\hline
\end{tabular}
}%
\caption{Forecasting performance for total admissions by ward. Values for XGBoost and LSTM represent the mean $\pm$ standard deviation across 10 independent experimental runs.}
\label{tab3:forecasting_performance_All}
\end{table}

Table \ref{tab3:forecasting_performance_All} summarises forecasting accuracy for total admissions by ward. In general, LSTM, SARIMAX and XGBoost deliver similar levels of performance across wards. Ward-level MAE values span roughly 0.6--7.3 patients per day, with typical errors of 1--4 patients, and differences between models within a ward are usually smaller than 0.3 patients. For Emergency Medicine, LSTM and XGBoost perform almost identically, while SARIMAX is clearly inferior. In General Medicine, Surgery and Paediatric SARIMAX attains the lowest MAE. Paediatric, Psychiatry and Cardiology remain difficult to predict for all three approaches, with MAPEs of about 45--64\%, reflecting high day-to-day variability. Neurology is an exception -- despite very similar MAEs, SARIMAX yields a substantially lower MAPE  than LSTM  and XGBoost, suggesting that it more accurately reproduces the small but systematic fluctuations in this low-volume ward. Comparing to the major complexity results, all three proposed models consistently outperform the seasonal naive baseline across every ward for total admissions.

At the aggregate level, predictive accuracy improves in relative terms. For total daily admissions, MAEs are $6.88 \pm 0.30$  for LSTM, $7.28$ for SARIMAX and $6.63 \pm 0.03$ for XGBoost, with corresponding MAPE values of $10.70\% \pm 0.27 \%$, $11.18\%$ and $10.32\% \pm 0.04 \% $. Thus, XGBoost offers the best overall performance for total ED admissions, although the performance gap between the three methods is small and each captures the principal temporal dynamics of the aggregate series.

\begin{figure}[h]
    \centering
    \includegraphics[width=\linewidth]{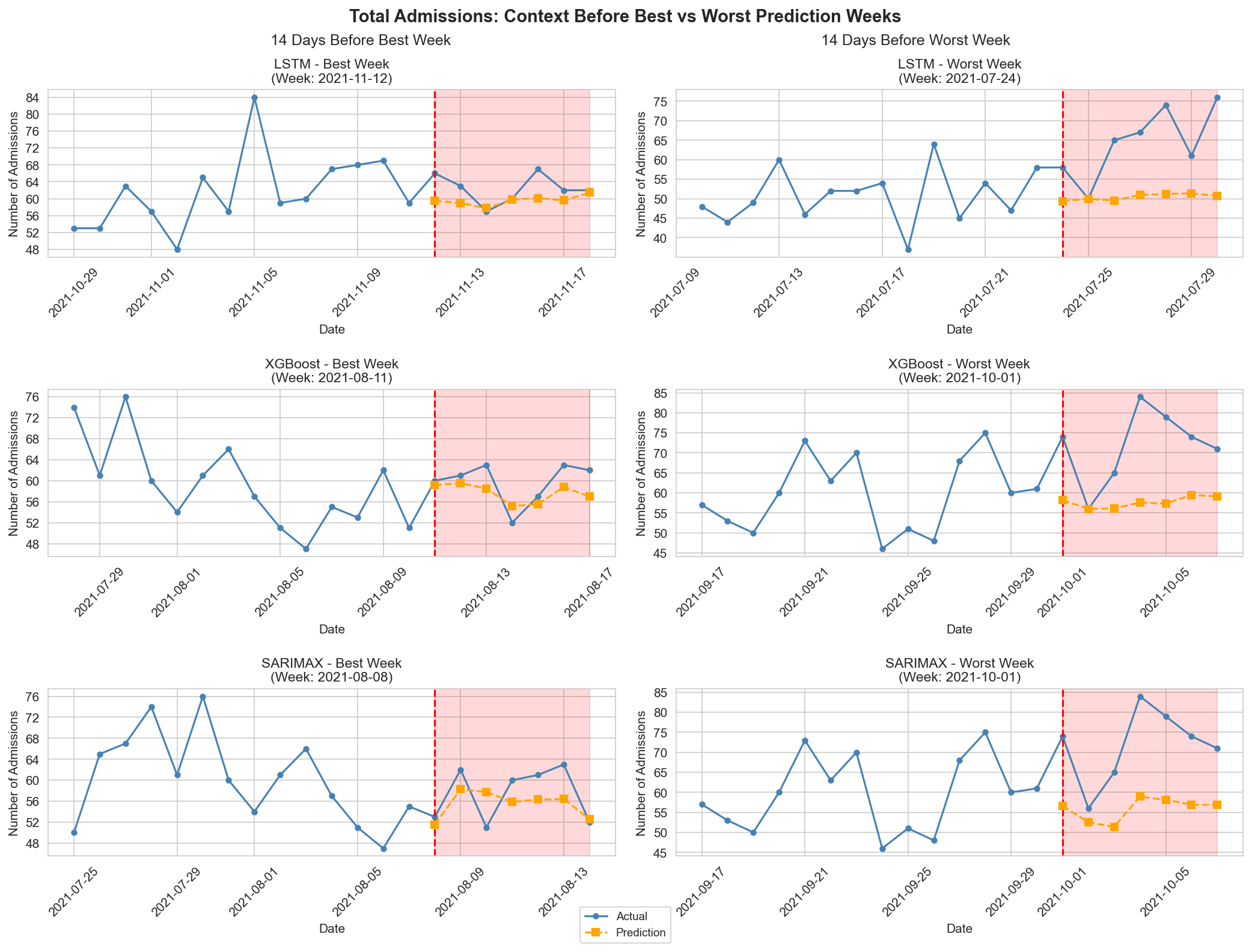}
    \caption{Actual and predicted total arrivals on best week and worst week -- comparison between XGBoost, LSTM and SARIMAX.}
    \label{best_week_all}
\end{figure}

Figure~\ref{best_week_all} provides the same diagnostic view for total daily admissions as in Figure~\ref{fig:best_model_predictions_comparison_major}. The graphs on the left confirm that when demand follows a standard weekly seasonality, all three models can generate highly accurate forecasts with low MAE. Among them, the XGBoost performs best with MAE $2.71$ followed by LSTM (MAE $3.14$), while SARIMAX yields a higher MAE $4.14$. In contrast, the worst week panels on the right correctly illustrate the \textit{outburst} phenomenon. In these weeks, all methods under-predict the observed counts and yield trajectories that are much flatter than the actual series. The LSTM attains the lowest error with MAE equals $14$, XGBoost performs slightly worse, MAE is $14.28$, and SARIMAX has the highest MAE equals to $16$.

\section{Conclusion}
\label{sec:conclusion}

The study demonstrates that advanced forecasting techniques are viable tools for predicting short-term ED patient demand. While XGBoost performed best for predicting total admissions and SARIMAX was marginally superior for major complexity cases, the performance gap between the statistical, machine learning, and deep learning approaches was narrow.

A critical finding is that while these models successfully reproduce regular day-to-day patterns, they share a common limitation in underestimating sudden, infrequent surges in patient volume. The study highlights the necessity of handling anomalous periods with synthetic data to maintain predictive accuracy. Ultimately, the reliance on autoregressive features for most wards suggests that recent historical trends remain the strongest indicator of near-future demand.

Furthermore, our ward-level decomposition revealed that high-volume departments, specifically Emergency Medicine, General Medicine, and Surgery, exhibited more stable demand patterns and were significantly easier to forecast, achieving lower relative errors than lower-volume wards like Paediatrics and Psychiatry.

Future work should prioritize addressing the models' limitation in capturing rare demand surges, potentially by exploring loss functions that penalize underestimation more heavily to better anticipate spikes. Given the struggle to predict low-volume wards like Neurology, investigating count-based statistical methods may offer superior handling of sparse data compared to the standard regression metrics used in this study. Finally, since SARIMAX excelled in predicting major complexity cases while XGBoost proved superior for total admissions, developing a weighted ensemble model that leverages the specific strengths of each architecture could significantly enhance overall predictive robustness.

\section{Statements and declarations}\label{Sec:Declarations}

\noindent{\bf Funding}\\
This work is partially supported by National Science Centre, Poland, under grant No. 2021/41/B/HS
4/00599.

\noindent{\bf Conflict of Interest}\\
The authors have no competing interests to declare that are relevant to the content of this article.

\noindent{\bf Data}\\
Data used in the paper was obtained following ethical approval from the  University of Tasmania Human Research (approval number H23633) and site-specific approval from the Research Governance Office of the Tasmanian Health Service.\\

\bibliographystyle{abbrv}
\bibliography{references_kuba}

\newpage

\appendix

\section{Feature importance and SHAP}
\label{app:feature_importance}

We present the model interpretability analysis for total admissions predictions. We provide the XGBoost global feature importance plot alongside local explanations derived from SHAP (SHapley Additive exPlanations). The SHAP analysis specifically targets the prediction sequences with the lowest and highest errors, as shown in Figure~\ref{best_week_all}. To summarize the multistep forecast, the SHAP values are aggregated to represent the mean feature contribution over the 7-day horizon.
\begin{figure}[h]
    \centering
    \includegraphics[width=\linewidth]{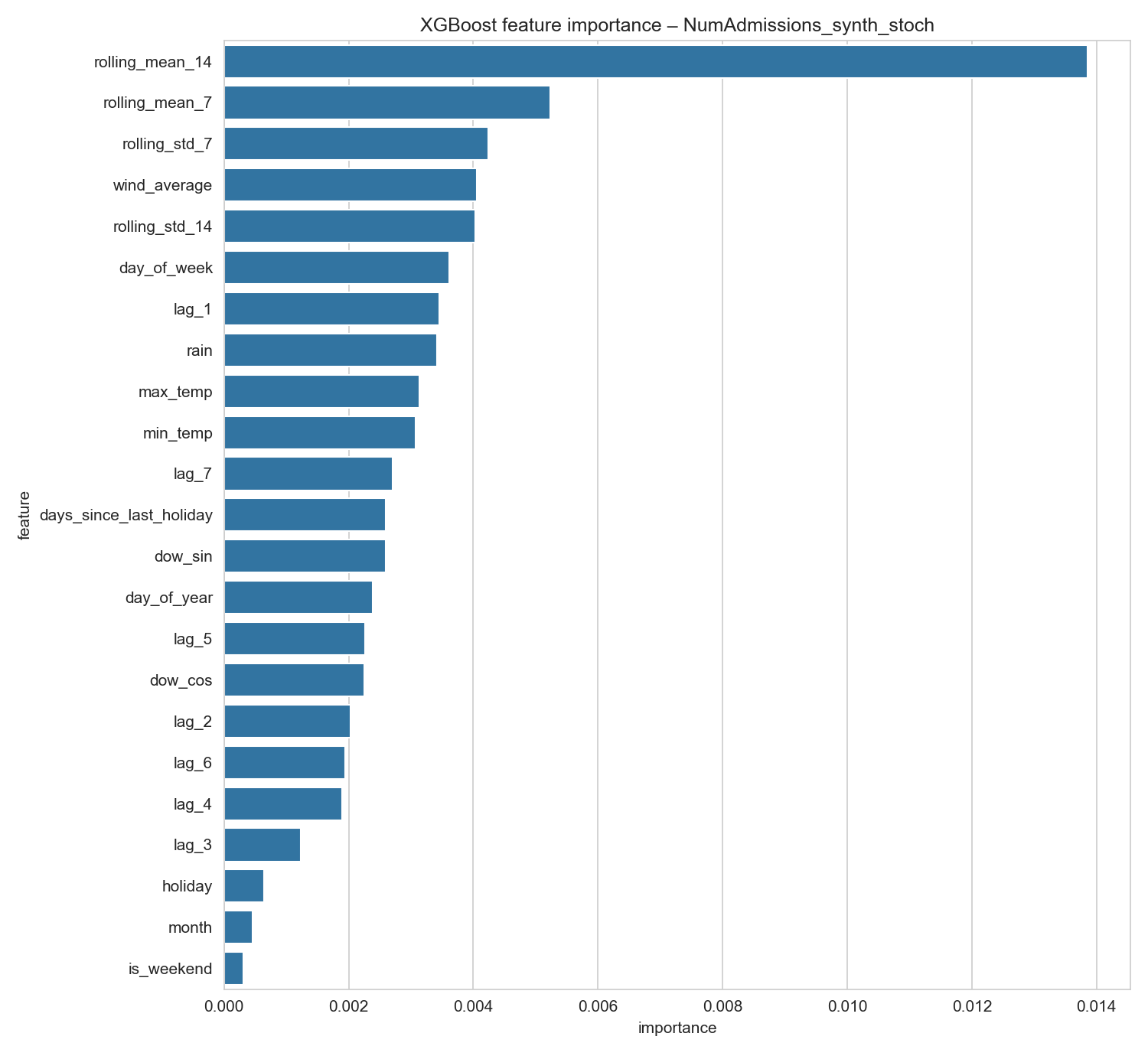}
    \caption{Feature importance for the XGBoost model predicting total hospital admissions.}
    \label{fig:feature_importance_total}
\end{figure}
\begin{figure}[h]
    \centering
    \begin{subfigure}[b]{0.48\textwidth}
            \includegraphics[width=\linewidth]{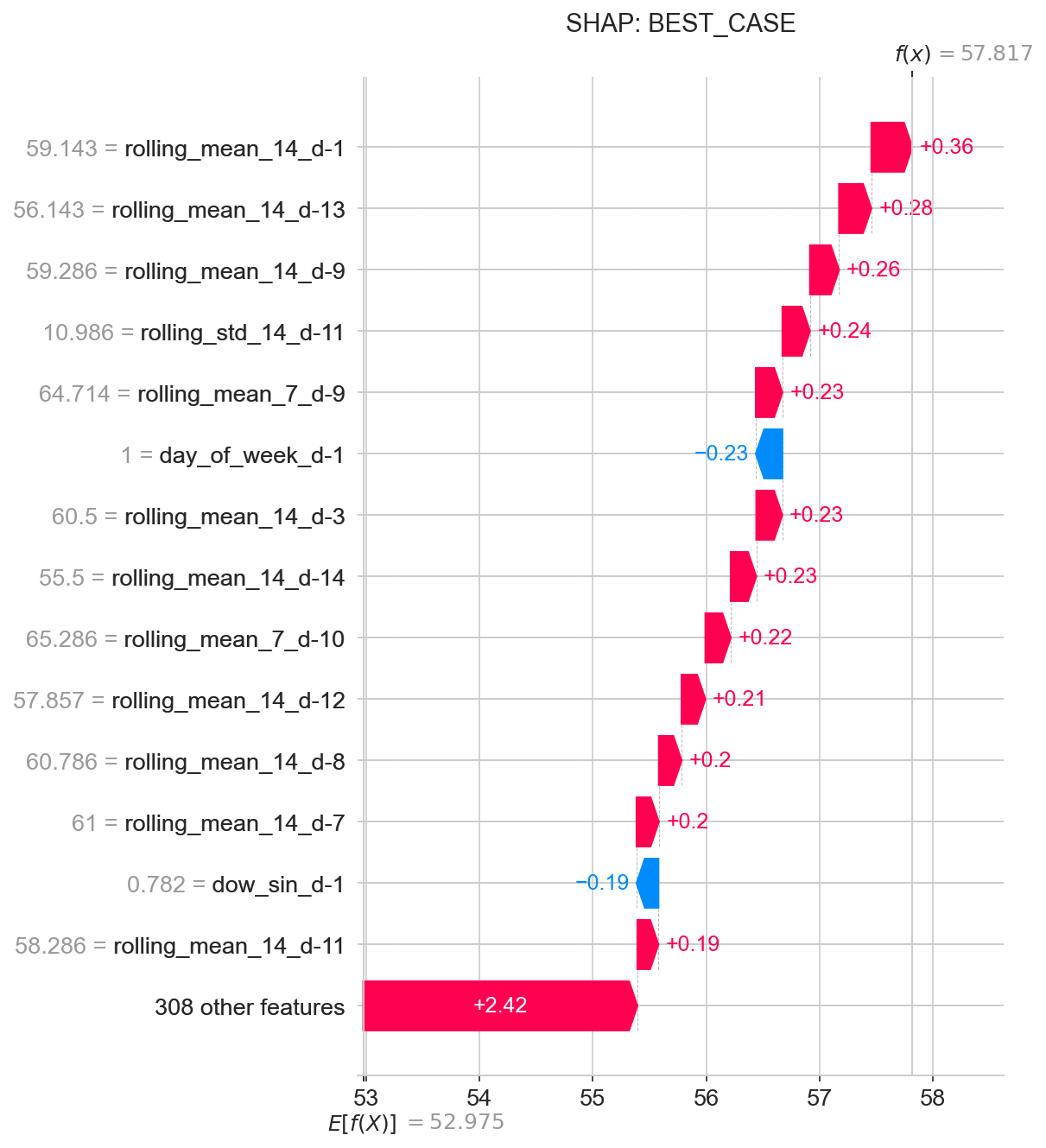}
            \caption{Best week}
            \label{fig:shap_best}
    \label{fig:shap_worst}
    \end{subfigure}
    \begin{subfigure}[b]{0.48\textwidth}
            \includegraphics[width=\linewidth]{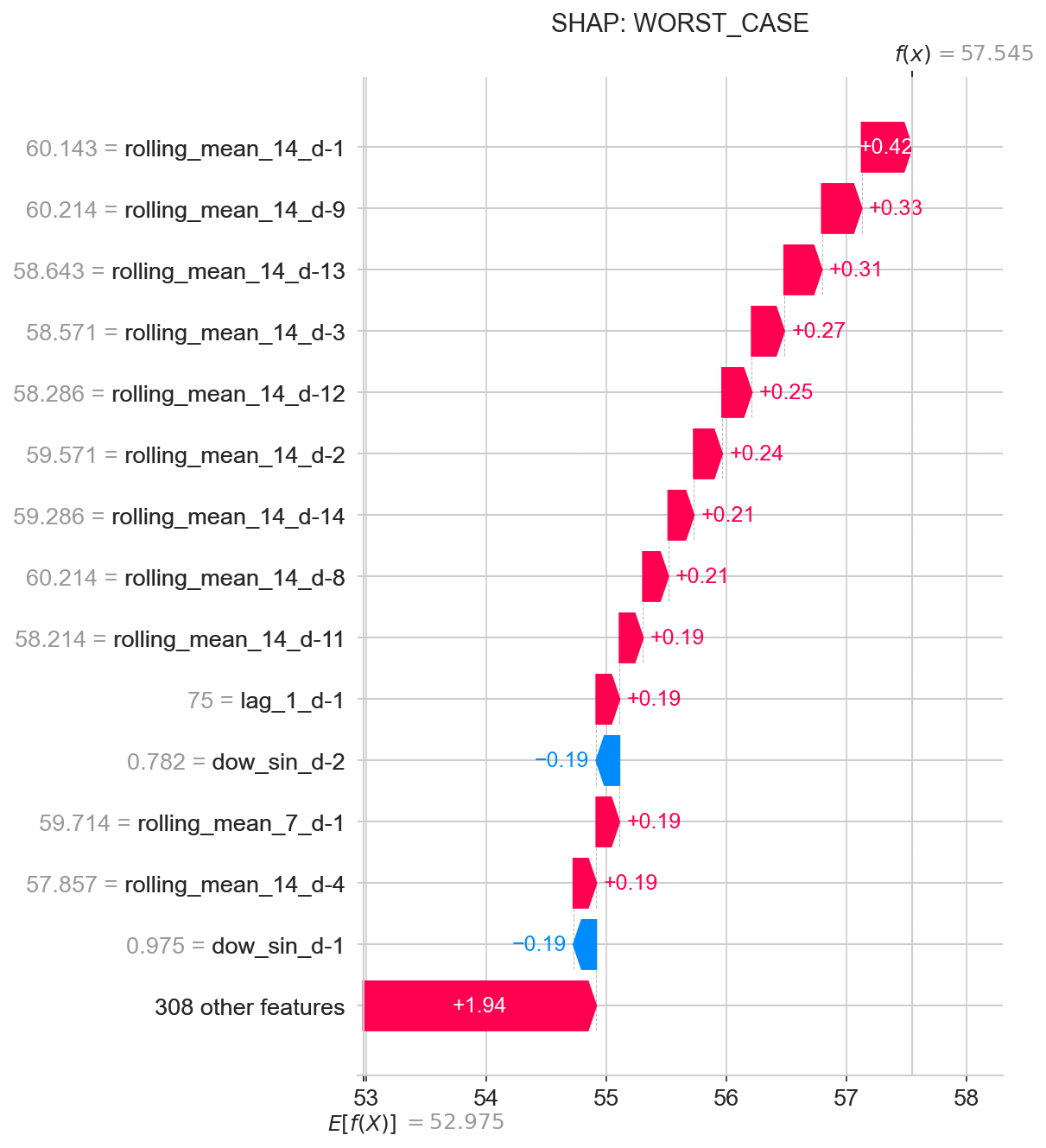}
            \caption{Worst week}
    \end{subfigure}
    \caption{SHAP waterfalls showing feature contribution to the weekly prediction for total admissions for different levels of prediction effectiveness.}
\end{figure}

\end{document}